\documentclass[nohyperref]{article}

\usepackage{microtype}
\usepackage{graphicx}
\usepackage{subfigure}
\usepackage{booktabs} 

\usepackage{hyperref}

\usepackage[accepted]{icml2022}

\usepackage{amsmath}
\usepackage{amssymb}
\usepackage{mathtools}
\usepackage{amsthm}
\usepackage{dsfont}

\usepackage[capitalize,noabbrev]{cleveref}

\theoremstyle{plain}

\theoremstyle{definition}

\theoremstyle{remark}

\usepackage{amsmath}
\DeclareMathOperator*{\argmax}{arg\,max}
\DeclareMathOperator*{\argmin}{arg\,min}
\usepackage[textsize=tiny]{todonotes}

\usepackage{xcolor}

\icmltitlerunning{Discovering Generalizable Spatial Goal Representations via Graph-based Active Reward Learning}

\begin{document}

\twocolumn[
\icmltitle{Discovering Generalizable Spatial Goal Representations via Graph-based Active Reward Learning}



\icmlsetsymbol{equal}{*}

\begin{icmlauthorlist}
\icmlauthor{Aviv Netanyahu}{equal,dept1}
\icmlauthor{Tianmin Shu}{equal,dept1,dept2}
\icmlauthor{Joshua Tenenbaum}{dept1,dept2}
\icmlauthor{Pulkit Agrawal}{dept1}
\end{icmlauthorlist}

\icmlaffiliation{dept1}{
Dept. of Electrical Engineering and Computer Science, Massachusetts Institute of Technology, Cambridge, MA}
\icmlaffiliation{dept2}{
Dept. of Brain and Cognitive Science, Massachusetts Institute of Technology, Cambridge, MA}

\icmlcorrespondingauthor{Aviv Netanyahu}{avivn@mit.edu}
\icmlcorrespondingauthor{Tianmin Shu}{tshu@mit.edu}

\icmlkeywords{Machine Learning, ICML}

\vskip 0.3in
]



\printAffiliationsAndNotice{\icmlEqualContribution} 

\begin{abstract}    

In this work, we consider one-shot imitation learning for object rearrangement tasks, where an AI agent needs to watch a single expert demonstration and learn to perform the same task in different environments. To achieve a strong generalization, the AI agent must infer the spatial goal specification for the task. However, there can be multiple goal specifications that fit the given demonstration. To address this, we propose a reward learning approach, Graph-based Equivalence Mappings (GEM), that can discover spatial goal representations that are aligned with the intended goal specification, enabling successful generalization in unseen environments. Specifically, GEM represents a spatial goal specification by a reward function conditioned on i) a graph indicating important spatial relationships between objects and ii) state equivalence mappings for each edge in the graph indicating invariant properties of the corresponding relationship. GEM combines inverse reinforcement learning and active reward learning to efficiently improve the reward function by utilizing the graph structure and domain randomization enabled by the equivalence mappings. We conducted experiments with simulated oracles and with human subjects. The results show that GEM can drastically improve the generalizability of the learned goal representations over strong baselines.\footnote{Project website: \url{https://www.tshu.io/GEM}}

\end{abstract}

\section{Introduction}
\label{intro}

To build AI agents that can assist humans in real-world settings, we have to first enable them to learn to perform any new tasks defined by a human user. To achieve this, an AI agent has to acquire two types of key abilities: i) the ability to develop an understanding of the goal or task specification intended by the human user and ii) the ability to execute a given goal. 
In this work, we aim to engineer the first key ability of an AI agent. 
This ability is crucial for a broad range of tasks such as housekeeping and manufacturing and is therefore commonly studied in robots \citep{shah2018bayesian, yan2020robotic, rowe2019desk} and embodied AI \citep{puig2021watchandhelp, srivastava2022behavior, kapelyukh2022my}.
As a foundation for these tasks, we focus on object rearrangement, where an agent must reason about the spatial goals that define a set of desired spatial relationships between objects. For instance, to set up a dinner table, one has to know how to place the plates and utensils appropriately.

\begin{figure}[t!]
\begin{center}
\centerline{\includegraphics[width=\columnwidth]{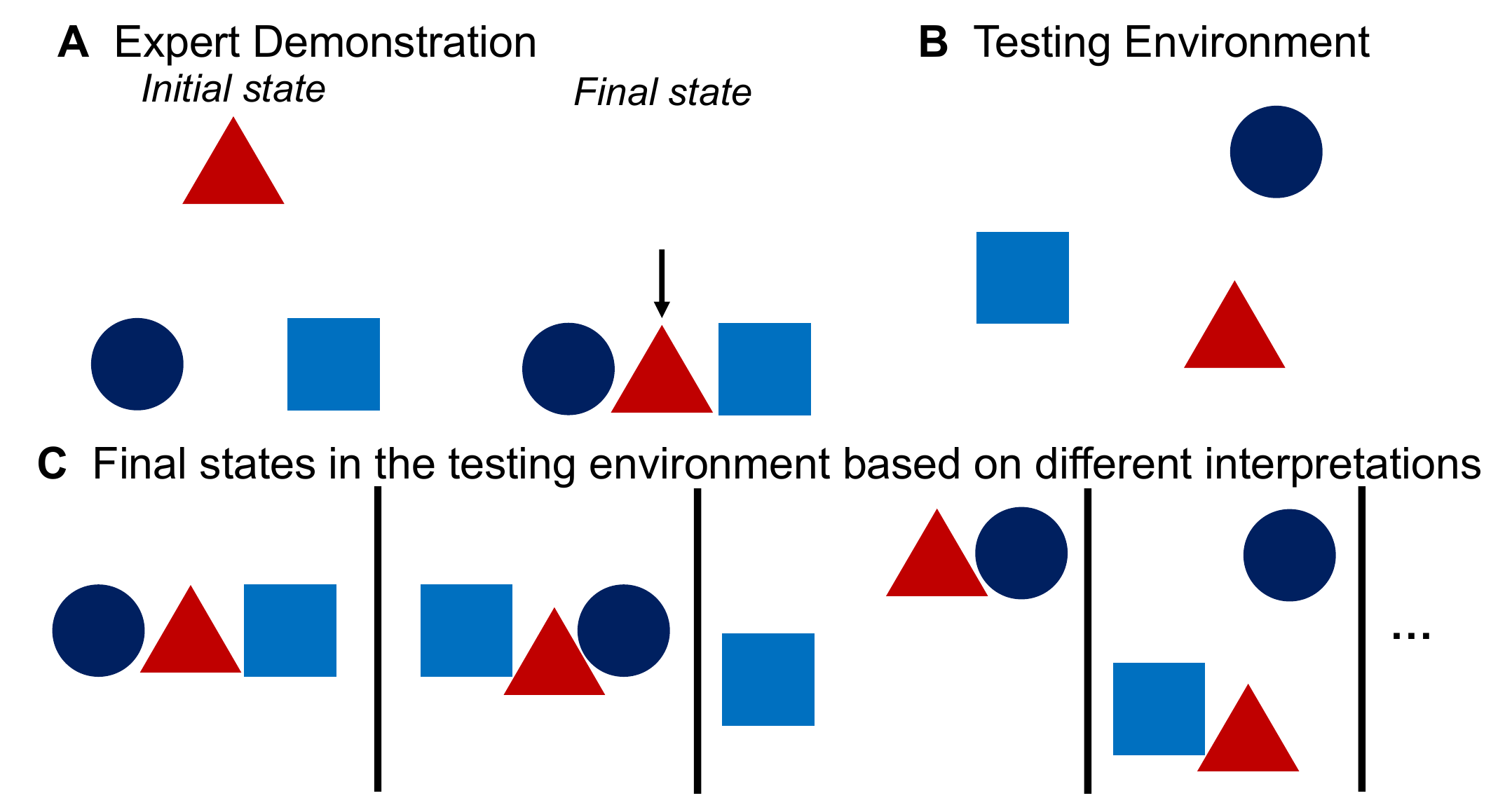}}
\caption{Illustration of our problem setup. We first (\textbf{A}) show a single expert demonstration for an object rearrangement task to an agent, and then (\textbf{B}) ask the agent to reach the same goal in unseen testing environments. (\textbf{C}) Multiple spatial goals can interpret the expert demonstration, each leading to a distinct task execution in the testing environments. For instance, from left to right, the four possible spatial goals shown here are i) triangle is to the right of circle and to the left of square; ii) triangle is close to circle and square; iii) triangle is to the left of circle; iv) triangle is close to square.} 
\label{fig:intro}
\end{center}
\vspace{-5pt}
\end{figure}

One way to train agents to perform a new object rearrangement task is to provide manual goal specification. However, creating a manual definition for the goal requires expert knowledge, and inaccurate definitions may cause misspecification \citep{amodei2016concrete}. 
Another widely used paradigm is imitation learning, in which an agent learns the task specification from demonstrations. Inferring the precise task specification from demonstrations typically requires collecting a large set of diverse demonstrations. However, diverse demonstrations are not always available in real-world applications due to the high cost of data collection. In this work, we instead learn from a single demonstration combined with queries to an oracle that can provide feedback over states.
Generally, in the one-shot imitation learning setup, the objective is to learn a new task from a single demonstration after pre-training on many tasks \citep{duan2017one}.
In our one-shot imitation learning setup, the agent watches the human user performing the task once (Figure~\ref{fig:intro}\textbf{A}) and learns via active learning to perform the same task in unseen environments (Figure~\ref{fig:intro}\textbf{B}) without pre-training on additional tasks.

While one-shot imitation learning is a convenient paradigm for teaching new tasks to agents, it is also extremely challenging due to the fact that there could be multiple goal specifications that explain the given expert demonstration well. For instance, consider the task illustrated in Figure~\ref{fig:intro}. There are multiple interpretations of the intended goal spatial relationships based on the demonstration, each of which will lead to a different task execution in a new environment (Figure~\ref{fig:intro}\textbf{C}). Without a correct understanding of the true goal, an agent cannot successfully perform the task.

To address this challenge, our work improves both i) the representation of spatial goal specification and ii) the acquisition of such representation that can reveal the true spatial goal.

\begin{figure}[t!]
\begin{center}
\centerline{\includegraphics[width=\columnwidth]{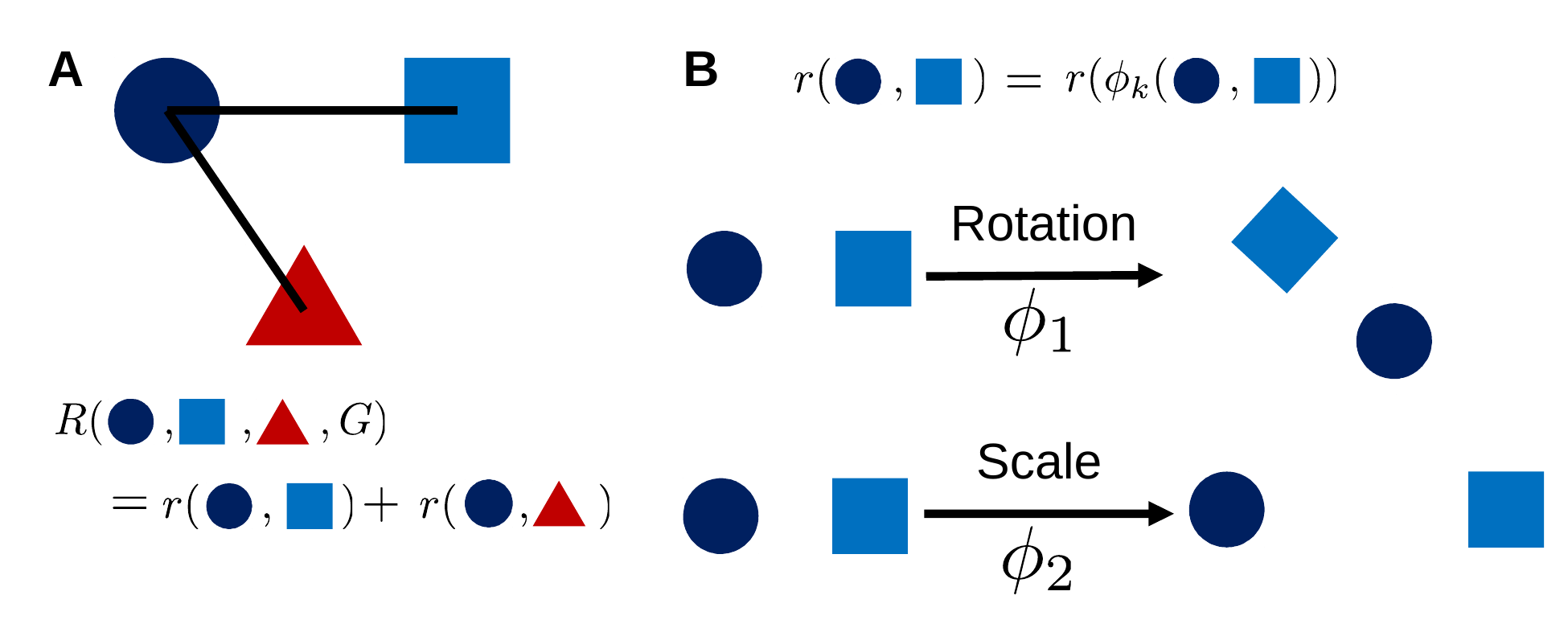}}
\caption{(\textbf{A}) We use a compositional reward function, $R$, conditioned on a graph, $G$, as the spatial goal representation. (\textbf{B}) For each edge, we may apply certain state equivalence mappings for improving representation learning to achieve a better generalization beyond the expert demonstration. Each type of mapping indicates a type of invariance of the intended spatial relationship between a pair of objects. Thus the reward for the edge would remain the same after applying the state mapping. Specifically, for the rotation-invariant mapping $\phi_1$, the relative orientation between objects can be randomized while the reward remains the same; for the scale-invariant mapping $\phi_2$, the change in the distance between objects does not affect the reward.}
\label{fig:GEM}
\end{center}
\vspace{-10pt}
\end{figure}

First, we represent the intended spatial goal by a compositional reward function conditioned on a sparse graph (Figure~\ref{fig:GEM}\textbf{A}) where the graph indicates \textit{whether} there is an important spatial relationship between a pair of objects and each edge has a reward function implicitly describing \textit{what} the intended spatial relationship is between the corresponding pair of objects. Unlike prior work on reasoning about spatial relationships and graphical representations of goals, we do not classify a relation out of a manually defined set of predicates (e.g., close, above), but intend to discover those predicates through the graph and the reward components for the edges in the graph implicitly.  

\begin{figure*}[t!]
\begin{center}
\centerline{\includegraphics[width=\textwidth]{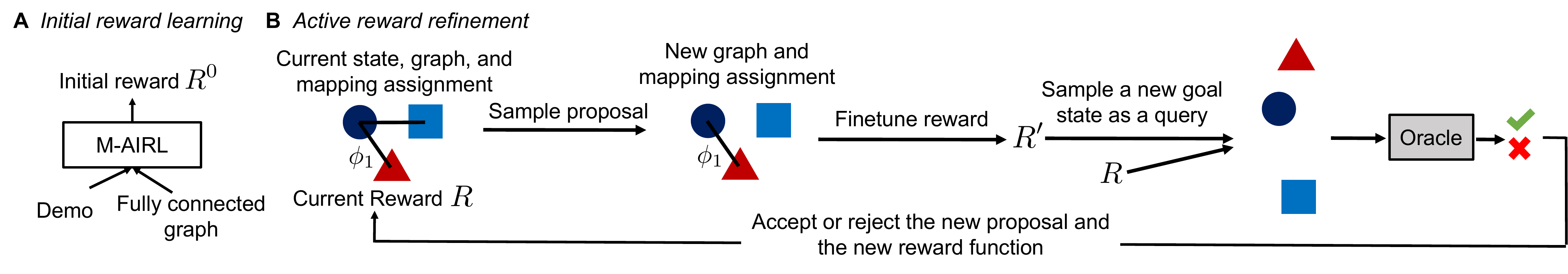}}
\caption{Overview of GEM. The reward learning consists of two phases, offering a novel connection between (\textbf{A}) model-based inverse reinforcement learning that predicts an initial reward and (\textbf{B}) active reward learning. Given an expert demonstration, we first initialize the reward function conditioned on a fully connected graph (i.e., $R^0$) using model-based adversarial inverse reinforcement learning (M-AIRL). $R^0$ provides a good estimation of expert demonstration state rewards with a theoretical fitness guarantee. To improve generalization beyond states in the expert demonstration, we update the reward function iteratively. At each iteration, we propose a new graph or a new equivalence mapping assignment for the edges in the new graph. We then finetune the reward function conditioned on the new graph using data augmented by the new equivalence mappings. To verify the fitness of the proposed graph and the equivalence mappings, we generate a new goal state that can differentiate the new reward $R^\prime$ from the current reward $R$ as an informative query for the oracle. Based on the oracle feedback (i.e., whether the new goal state is acceptable), we update the current proposal, reward function, and state accordingly. We also collect the query states as additional training data for the reward finetuning at future iterations.}
\label{fig:overview}
\end{center}
\end{figure*}

Second, we propose a novel reward learning algorithm, Graph-based Equivalence Mappings (GEM), connecting offline reward learning with active reward learning. As shown in Figure~\ref{fig:overview}, GEM consists of two phases. In the initial phase (Figure~\ref{fig:overview}\textbf{A}), we first learn a reward function conditioned on a fully connected graph from the expert demonstration using adversarial inverse reward learning (AIRL) \cite{fu2017learning} with a model-based extension, M-AIRL. This reward function is guaranteed to provide a good fit on the states in the expert demonstration, which offers us a set of good state-reward pairs as a training set. However, this set is only limited to the training situation since M-AIRL penalizes states that were not in the demonstration. Thus the initial reward function learned from M-AIRL may not generalize well to states that were not seen in training. To acquire new state-reward pairs that are not covered by the expert demonstration and improve the spatial goal representations to match with the newly acquired data, we then conduct active reward refinement (Figure~\ref{fig:overview}\textbf{B}) following the initial training.

Prior works on active reward learning typically collect new state-reward training data purely based on the states generated in the queries shown to the oracle and the feedback received from the oracle. When faced with a large state space, this paradigm often requires a large number of queries to acquire sufficient training data.
To overcome this, we utilize spatial relation invariance and reward graph structure, resulting in a more efficient querying process.
Specifically, we aim to augment the existing training states by randomizing each state in the existing set without changing the corresponding reward. This can be achieved by applying state equivalence mappings (Figure~\ref{fig:GEM}\textbf{B}) to edges, which is a type of state transformation that identifies equivalent states. A similar idea has been previously applied to multi-agent policy learning for improving zero-shot generalization \cite{hu2020other}. Here, each type of equivalence mapping indicates a type of invariance for the intended spatial representation between a pair of objects. Thus, when applying valid equivalence mappings, we can synthesize new state-reward pairs with known rewards without the need to acquire oracle feedback on each new state. Following this intuition, our active reward refinement iteratively proposes a new hypothesis including a new graph and new state equivalence mappings assigned to the edges in the graph, finetunes the reward function based on the new graph and the augmented training set, and generates informative queries for the oracle to verify the hypothesis. For instance, in Figure~\ref{fig:overview}\textbf{B}, the edge between the square and the circle is removed in the new graph, while the rotation-invariant mapping is preserved for the remaining edge. Consequently, the sampled query moves the square away and rotates the triangle around the circle, so that we can verify whether the removed edge is important and whether the rotation-invariance holds for the connected edge through this query.

We conducted experiments with a simulated oracle and with human subjects in a 2D physics simulation environment, Watch\&Move. In each task, the goal is to move the objects to satisfy spatial relationships intended by the oracle. We compared GEM against recent baselines for imitation learning and active reward learning and found that GEM significantly outperformed the baselines, both in terms of generalizability and sample efficiency (i.e., fewer oracle queries).

In summary, our main contributions are: i) a generalizable spatial goal representation using a compositional reward function conditioned on a graph and state equivalence mappings, ii) a novel reward learning algorithm, GEM, for discovering spatial goal representations by connecting inverse reinforcement learning with sample-efficient active reward learning, and iii) a new physics simulation environment, Watch\&Move, for evaluating one-shot imitation learning approaches, with a focus on generalization.

\section{Related Work}

\noindent{\textbf{Goal inference}}. One of the key aspects of the work is to reason about the goal for a task based on the expert's plan. There has been rich history in goal inference building socially intelligent AI systems \cite{baker2017rational, puig2021watchandhelp, NetanyahuPHASE2021, shah2018bayesian, yan2020robotic}.
However, prior work on goal inference typically assumed a limited goal space such as a set of discrete goals \cite{baker2017rational}, a finite set of predicates \cite{puig2021watchandhelp, NetanyahuPHASE2021, shah2018bayesian}, or a target location \cite{cao2020long}. These assumptions greatly limited the kinds of goals a system can infer. In contrast, our approach can discover generalizable spatial goal representations with much less restriction for the spatial goal specification.

\noindent{\textbf{Inverse reinforcement learning}}. Most existing imitation learning approaches can be categorized into behavioral cloning (BC) and inverse reinforcement learning (IRL) \cite{ghasemipour2020divergence}. These two types of methodologies provide two distinct learning objectives -- BC aims to directly mimic the expert policy, whereas IRL attempts to recover the reward function that could produce the expert policy. 
Intuitively, learning a reward can achieve better generalization in novel environments since the learned reward function may still be valid in the new environment, whereas a policy inferred by demonstrations may no longer be suitable when the environment distribution changes (covariate shift) \cite{shimodaira2000improving}.
The main challenge in IRL is that there are usually multiple rewards that can explain the expert demonstrations \cite{ng2000algorithms}, especially with limited demonstrations.
Policies learned from IRL simultaneously with the reward are guaranteed to only perform well on the expert distribution \cite{ghasemipour2020divergence}. 
State-only AIRL \cite{fu2017learning} is guaranteed to learn a reward disentangled from environment dynamics, but may also suffer from covariate shift. Hence, we propose a new learning paradigm that extends IRL with graph-based active reward learning.

\noindent{\textbf{One-shot imitation learning}}. 
Imitation learning has long been a subject of interest \cite{schaal1999imitation,nehaniv2002correspondence,abbeel2004apprenticeship, billard2004discovering,argall2009survey}. Specifically, there has been work on one-shot imitation learning \cite{duan2017one,bonardi2020learning,huang2019continuous,yu2018one}, which often adopted a meta-learning framework, where the objective is to learn how to learn from a single demonstration through training with a distribution of tasks. This can be done either through meta policy learning \cite{finn2017one} or meta reward learning \cite{xu2019learning}. However, existing works focused on tasks that have a high similarity, e.g., pushing an object to different locations \cite{finn2017one}. They have also not addressed the generalizability of the learned policies or reward functions to unseen environments. There are two main challenges for achieving a strong generalization in one-shot imitation learning: it is hard i) to learn to reach a given goal in unseen environments and ii) to infer the true goal from a single demonstration. In this work, we focus on the second challenge, with the assumption of having access to a world model and a planner that can reach any physically plausible goal state. We believe this is a first step towards engineering a generalizable one-shot imitation learning system.

\noindent{\textbf{Policy/reward learning from human feedback}}. In addition to learning from demonstrations, prior work has also proposed methods for learning policies \cite{ross2011reduction,griffith2013policy,loftin2014strategy,macglashan2017interactive,arumugam2019deep,wang2022skill}, or reward functions \cite{daniel2014active,daniel2015active,su2016line,biyik2019asking,cui2018active,brown2019extrapolating,reddy2020learning} from human feedback. This can be achieved through queries that ask for human preferences \cite{christiano2017deep, brown2019extrapolating} or a direct evaluation (reward) on states \cite{ross2011reduction,reddy2020learning}. Inspired by this, our approach also uses an active reward learning scheme to improve the reward function from oracle feedback. However, when the state and action spaces are large, it is difficult to obtain a sufficient amount of data from a small number of queries. In this work, we aim to address this by better utilizing the limited queries and oracle feedback. Specifically, instead of only generating trajectories or states for the queries, we propose generalizable goal representations and verify them through smart query generation.

\section{Preliminaries} \label{sec:AIRL}

Our initial learning phase adopts adversarial inverse reinforcement learning (AIRL) \cite{fu2017learning}, which was proposed to achieve robust reward generalization for unseen dynamics. We briefly introduce AIRL and present a model-based extension to the original AIRL below.

\subsection{Adversarial Inverse Reinforcement Learning}

IRL considers an MDP process $\langle \mathcal{S}, \mathcal{A}, \mathcal{T}, r, \gamma, \rho_0 \rangle$. $\mathcal{S}$ is the state space, $\mathcal{A}$ is the action space, $\mathcal{T}(\cdot|a,s)$ is the state transition distribution, $r(s,a)$ is the reward function, $\gamma$ is the discount factor, and $\rho_0$ is the initial state distribution. The goal of IRL is to learn a reward function $r_\theta(s, a)$ that can approximate the expert policy on the given expert demonstrations $\mathcal{D}=\{\Gamma_1, \cdots, \Gamma_M\}$, where $\Gamma_i = \{(s^t, a^t)\}_{t=0}^T$ is a sequence of state and action pairs in a demonstration. It achieves this objective by maximizing the likelihood of observing the expert demonstrations given the reward function.
\begin{equation}
\max_\theta E_{\Gamma \sim \mathcal{D}}[\log p_\theta(\Gamma)],
\end{equation}
where $p_\theta(\Gamma)\propto p(s_0)\prod_{t=0}^T p(s^{t+1}|s^t, a^t)e^{\gamma^t r_\theta(s^t, a^t)}$ is the likelihood of the demonstrations given the reward function. AIRL formulates this optimization as adversarial training, where it learns to approximate the advantage function for the expert policy through a discriminator. The discriminator distinguishes between generated trajectories from a learned policy $\pi(a |s)$ (as fake examples) and the expert demonstrations (as real examples):
\begin{equation}
   D_{\theta,\omega}(s, a, s^\prime) = \frac{\exp\{f_{\theta,\omega}(s, a, s^\prime)\}}{\exp\{f_{\theta,\omega}(s, a, s^\prime)\} + \pi(a | s)},
\label{eq:discrim}
\end{equation}
where $f_{\theta,\omega}$ is the advantage function consisting of an approximated reward function $g_\theta$ as well as a shaping function $h_\omega$:
\begin{equation}
    f_{\theta,\omega} = g_\theta(s,a) + \gamma h_\omega(s^\prime)-h_\omega(s).
    \label{eq:advantage_func}
\end{equation}
When the reward function only depends on state $s$, AIRL can guarantee that the learned reward function $g_\theta$ and the shaping function $h_\omega$ can approximate the ground truth reward function $r^*$ and the ground truth value function $V^*(s)$ up to a constant respectively. However, the learned reward may not generalize well to states different from the ones shown in the expert demonstrations.

\subsection{Model-based AIRL} \label{sec:M-AIRL}
The original AIRL uses model-free RL training, which is often difficult in tasks with large state and action spaces (such as the multi-object rearrangement tasks studied in this work).
To address this issue there has been work on model-based AIRL \cite{sun2021adversarial}. Similarly, we propose a simple model-based extension given a world model $p(\cdot|s,a)$ learned or given by a simulator. The idea is to utilize the shaping function $h(s)$ in Eq.~(\ref{eq:advantage_func}), which is guaranteed to approximate the ground truth value function in the demonstrations, combined with one-step-ahead state predication to define the policy in Eq.~(\ref{eq:discrim}) instead of learning a policy $\pi$ using model-free RL:
\begin{equation}
    \pi(a|s)= \frac{\exp\{\beta \sum_{s^\prime}p(s^\prime|a,s)h(s^\prime)\}}{\sum_{a^\prime\in \mathcal{A}}\exp\{\beta \sum_{s^\prime}p(s^\prime|a^\prime,s)h(s^\prime)\}},
    \label{eq:model-based-policy}
\end{equation}
where $\beta$ is a constant coefficient.

\section{Graph-based Equivalence Mappings}

As a representation of a spatial goal, a reward function can adequately describe the fitness of a spatial configuration w.r.t. any intended goal spatial relationships including both logical and continuous relationships. When learned properly, the trained reward function also enables a generalization of the corresponding goal in unseen physical environments, avoiding over-imitation, unlike direct policy imitation. However, without a diverse set of demonstrations, multiple reward functions may perfectly fit the expert trajectories, and it is impossible to disambiguate the true reward solely based on the given demonstration. To solve this, we propose a novel active reward learning approach, Graph-based Equivalence Mappings (GEM) that learns a compositional reward function conditioned on a sparse graph and state equivalence mappings (Figure~\ref{fig:GEM}). As illustrated in Figure~\ref{fig:overview}, GEM consists of two learning phases, combining both model-based inverse RL and active reward learning. Starting from the initial reward function that has a theoretical fitness guarantee limited to the expert demonstration, GEM iteratively refines the reward function through proposed new graphs and state equivalence mappings and verifies the new reward function via informative queries for an oracle. In this section, we first introduce our compositional reward function and then present the two-phase learning algorithm.

\subsection{Compositional Reward Function as a Spatial Goal Representation}

We represent each state as $s=(x_i)_{i\in N}$, where $N$ is a set of objects, and $x_i$ is the state of object $i$. We indicate the important spatial relationships for a task by a graph, $G=(N, E)$, where each edge $(i, j) \in E$ shows that the spatial relationship between object $i$ and object $j$ is part of the goal specification. Here, we focus on pairwise spatial relationships, but it is possible to extend this to higher-order relationships. As shown in Figure~\ref{fig:GEM}\textbf{A}, given the graph and the state, we define a compositional reward function as a spatial goal representation to implicitly describe the goal spatial relationships for a task:
\begin{equation}
    R(s, G) = \frac{1}{|E|}\sum_{(i, j) \in E} r(x_i, x_j).
\end{equation}

Spatial relationships may have certain invariance properties. For instance, relationships describing the desired distance between two objects are invariant to the rotation applied to this pair of objects. By utilizing correct invariance properties, we can transform a state seen in the training environment to a new state that has the same reward, as they represent the same spatial relationship. Essentially this process augments the data by domain randomization \cite{tobin2017domain}. To model different invariance properties, we introduce a set of possible state equivalence mappings $\{\phi_k\}_{k\in K}$ as shown in Figure~\ref{fig:GEM}\textbf{B}, where each type of mapping $\phi_k$ can transform the states of a pair of objects $(i, j)$, and ensures that the reward component for that edge does not change, i.e., $r(x_i, x_j) = r(\phi_k(x_i, x_j))$. When applying a mapping, we may randomize the invariance aspect of the state to sample a new state. For instance, for applying the rotation-invariant mapping once, we randomize the relative orientation between the two objects for the state transformation. Please refer to Appendix~\ref{sec:app_mappings} for more details.

We denote the mappings assignment for all the edges in a graph as $I=\{\delta_{i,j,k}\}_{(i,j)\in E, k\in K}$, where $\delta_{i,j,k}$ is a binary variable indicating whether $\phi_k$ can be applied to edge $(i,j)$. The state mapping for the whole graph is then defined as $\Phi(s, I)$, where it recursively applies mappings assigned to each edge. Note that we transform the state for each edge independently so that the transformation of one edge will not affect the state of other edges that share a common node with this edge. For more details, please refer to Appendix~\ref{sec:app_mappings}. 

In this work, we consider two types of mappings illustrated in Figure~\ref{fig:GEM}\textbf{B} as a gentle inductive bias provided by domain knowledge, but other types of mappings can also be applicable to different domains. Critically, we only provide a set of candidate mappings but do not assume to have the knowledge of which mappings can be applied to a specific spatial relationship, unlike \cite{hu2020other}. We instead learn to assign valid mappings to each edge through active reward learning. 

These equivalence mappings allow us to augment the states from the expert demonstration and those acquired through queries, creating an infinite number of states that have equivalent spatial relationships.

\subsection{Two-Phase Reward Learning}

\subsubsection{Initial Reward Learning}

Given the expert demonstration $\Gamma$, we first use M-AIRL described in Section~\ref{sec:AIRL} to train an initial reward function $R^0(s, G^0;\theta^0)$ conditioned on a fully connected graph $G^0$, where $\theta^0$ are the parameters of the initial reward function. This initial reward function provides a good reward approximation for all the states in $\Gamma$, approaching the ground truth with a constant offset.

\begin{algorithm}[t!]
   \caption{Active Reward Refinement}
   \label{alg:active_learning}
\begin{algorithmic}
   \STATE {\bfseries Input:} $G_0$, $I_0$, $R_0$, $S_D$, $L_\text{max}$
   \STATE $S_+ \leftarrow \{s^T\}$, $S_- \leftarrow \emptyset$, $I_0 \leftarrow \emptyset$
   \STATE First reach the final state of the expert demonstration, denoted as $s_q^0$
   \FOR{$i=1$ {\bfseries to} $L_\text{max}$}
    \STATE $G^\prime, I^\prime  \sim Q(G^l, I^l | G^{l-1}, I^l)$
    \STATE Train $\theta^\prime$ based on Eq.~(\ref{eq:optimization})
    \STATE Sample a new query state $s_q^\prime$ based on Eq.~(\ref{eq:query_gen})
    \STATE Reach $s_q^\prime$ and get oracle feedback $o$
   \IF{$o = \text{acceptable}$}
   \STATE $G^l \leftarrow G^\prime$, $I^l \leftarrow I^\prime$, $s_q^l \leftarrow s_q^\prime$, $\theta^l \leftarrow \theta^\prime$
   \STATE $S_+ \leftarrow S_+ \cup \{s_q^\prime\}$, $S_D \leftarrow (s_q^\prime, r^T)$
   \ELSE
   \STATE $G^l \leftarrow G^{l-1}$, $I^l \leftarrow I^{l-1}$, $s_q^l \leftarrow s_q^{l}$, $\theta^{l-1} \leftarrow \theta^{l-1}$
   \STATE $S_- \leftarrow S_- \cup \{s_q^\prime\}$
   \STATE Move objects back to $s_q^{l-1}$
   \ENDIF
   \ENDFOR
\end{algorithmic}
\end{algorithm}

\subsubsection{Active Reward Refinement}

To improve the generalization of the initial reward function to states beyond expert states, we then aim to discover a graph structure, $G$, and equivalence mapping assignment for the edges in the graph, $I$, through active learning, and refine the reward function based on the inferred graph and the equivalence mappings. 

The active reward refinement is outlined in Algorithm~\ref{alg:active_learning}. For the reward refinement, we consider three training sets, (1) all states in the expert demonstrations and their rewards based on $R^0$, $S_D = \{(s^t, r^t=R^0(s^t|G^0,\theta^0))\}_{t=0}^T$, (2) a positive state set initialized with the final state of the expert demonstration $S_+$, and (3) a negative state set, $S_-$. By querying, we collect more states for the positive set and the negative set based on the oracle judgment. We know the corresponding reward of the states in $S_D$ based on $R_0$, and assume that rewards for states in $S_+$ have higher rewards than any state in $S_-$.

The main purpose of the queries is to find equivalently good goal states that are visually different from the goal state shown in the expert demonstration. To this end, at the start of the active learning phase, we first reach the final state of the expert demonstration. We can achieve this by applying the learned approximated value function (the policy is defined in Eq.~(\ref{eq:model-based-policy})).

At each iteration $l$, we first propose a new graph $G^\prime$ and a new equivalence mapping assignment $I^\prime$ based on a proposal distribution $Q(G^\prime, I^\prime | G^{l-1}, I^l)$. We describe the design of this proposal distribution in Appendix~\ref{sec:app_proposal}. We then finetune the reward function conditioned on the proposal to obtain new parameters $\theta^\prime$, which defines a new reward function $R(s|G^\prime;\theta^\prime)$. To do that, we use two types of optimization. First is reward regression based on the state and reward pairs $(s, r)\in S_D$. Since we obtained equivalence states from the proposed mappings, the reward function can be optimized so that for each state $s$ in $S_D$, all of its equivalent states will have the same reward as $s$ itself. We formally define a regression loss as follows
\begin{equation}
    \mathcal{L(\theta)}_\text{reg} = \mathbb{E}_{(s,r)\sim S_D}[(R(\Phi(s, I^\prime)|G^\prime;\theta)-r)^2].
\end{equation}
The second type of optimization is reward ranking. Specifically, we optimize the reward function so that the reward of a state $s_+ \in S_+$ is higher than the reward of any state $s_- \in S_-$. This gives us the second loss:
\begin{align*}
    \mathcal{L(\theta)}_\text{rank}  
    = \mathbb{E}_{s_+\sim S_+, s_-\sim  S_-}[|(R(\Phi(s_-, I^\prime)|G^\prime;\theta)
     - \\R(\Phi(s_+, I^\prime)|G^\prime;\theta)|_+].
\end{align*}
We then combine these two loss functions to update the parameters of the reward function:
\begin{equation}
    \theta^\prime = \argmin_\theta \mathcal{L(\theta)}_\text{reg} + \mathcal{L(\theta)}_\text{rank}.
    \label{eq:optimization}
\end{equation}

Based on the new reward, we generate a query to reflect the change in the hypothesis and in the corresponding reward function. The query is a goal state $s_q^\prime$ sampled starting from the current state (i.e., $s_q^{l-1}$). Intuitively, this new state should have a high reward based on the new reward function but a low reward based on the previous reward function at iteration $l-1$. Formally, $s_q^\prime$ is sampled by
\begin{equation}
    s_q^\prime = \argmax_{s\in \mathcal{N}(s_q^{l-1})} R(s|G^\prime;\theta^\prime) - \lambda  R(s|G^{l-1};\theta^{l-1}),
    \label{eq:query_gen}
\end{equation}
where $\mathcal{N}(s_q^{l-1})$ is the set of all reachable states starting from $s_q^{l-1}$ and $\lambda$ is a constant coefficient.

After reaching the new query state $s_q^\prime$, we query for oracle feedback by asking if $s_q^l$ is an acceptable state that satisfies the goal. If acceptable, we then accept the new proposal as well as the new reward function, and augment $S_+$ with the new query $s_q^\prime$. We also assume that $s_q^\prime$ has a similar reward as the final demonstration state. Thus $S_D$ can also be augmented with $(s_q^\prime, r^T)$, where $r^T$ is the reward of the final demonstration state according to the initial reward function. If the oracle feedback is negative, then we reject the new proposal and the corresponding reward function, move back to the last accepted state (i.e., $s_q^{l-1}$), and augment $S_-$ with $s_q^\prime$. We repeat this process until reaching the maximum number of iterations $L_\text{max}$. Alternatively, we can also terminate the process when no sparser graphs have been accepted for a certain amount of iterations. 

After the two learning phases, we apply the learned reward function to the test environments by sampling a goal state based on the reward function. Since we prefer sparse graphs, we use the sparsest graphs accepted in the recent iterations. When there are multiple accepted graphs with the same number of edges, we use the one that has the most mappings assigned to its edges.

\begin{figure*}[t!]
\begin{center}
\centerline{\includegraphics[width=\textwidth]{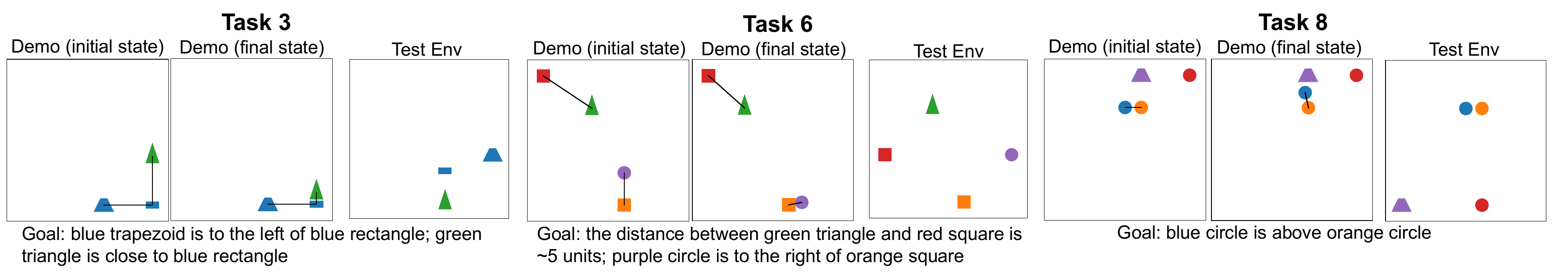}}
\caption{Example Watch\&Move tasks. For each task, we create an expert demonstration that moves objects to a state that satisfies the goal and provide a test environment that is different from the initial state in the expert demonstration. The edges visualize the relationships that are part of the goal definition; they are not present in the expert demonstration and are thus unknown to the agent. We also show the goal for each task. The distance in Watch\&Move is measured in units. As a reference, the radius of the circle is 0.8 units. The exact definitions of the goal relationships are summarized in Table \ref{tab:predicate-def} (Appendix~\ref{sec:app_watchandmove}).}
\label{fig:example_tasks}
\end{center}
\end{figure*}

\section{Experiments}

\subsection{Watch\&Move Environment}\label{sec:watchandmove}
We propose a one-shot imitation learning environment, Watch\&Move. Inspired by recent machine learning benchmarks based on 2D physics engines \cite{lowe2017multi, bakhtin2019phyre, allen2020rapid, NetanyahuPHASE2021}, we create a 2D physics simulation for Watch\&Move, where objects can be moved w.r.t. physics dynamics and constraints. This simulation environment provides an abstraction of real-world object rearrangement tasks\footnote{We illustrate this in Appendix~\ref{sec:app_vh}.}.

We design 9 object rearrangement tasks in the Watch\&Move environment (Figure~\ref{fig:example_tasks} shows 3 example tasks). Each task consists of an expert demonstration and a new testing environment. The goals of the proposed tasks cover a range of object relationships that are common in the real world as well as in the prior work on object rearrangement tasks (as summarized in Table~\ref{tab:predicate-def}).

Successful performance in the testing environment in Watch\&Move requires an agent to rearrange the objects to satisfy all spatial goal relationships while minimizing the overall change in the environment. This requires that the agent correctly identify the necessary relationships for each task. We use a reward function to measure the task completion and the displacement of all objects: $R_\text{eval} = \mathds{1}(s^T\text{ satisfies the goal}) - 0.02 \sum_{i \in N} ||x_i^0-x_i^T||_2$, where $x_i^0$ and $x_i^T$ are the first and last states of object $i$ in a testing episode respectively, and $s^T$ is the overall final state in the testing episode. Please refer to Appendix~\ref{sec:app_watchandmove} for more details about the tasks and the environment.

\subsection{Baselines and Ablations}

We compare GEM with M-AIRL (i.e., the initial reward learning alone) and with ReQueST \cite{reddy2020learning}, a recent active reward learning approach that estimates a reward that can generalize to environments with different initial state distributions. For a fair comparison, we use the same graph-based reward function for ReQueST and initialize the reward for using the expert demonstration. For details, please refer to Appendix~\ref{app:request}. For the ablated study, we also evaluate variants of GEM, including i) GEM without minimizing the previous reward for the query generation, ii) GEM with a fully connected graph (no new graph proposals), iii) GEM without applying any equivalence mappings, and iv) GEM trained with randomly generated queries. During testing, given the reward function learned by each method, we sample a goal state by maximizing the reward and minimizing the displacement. We then evaluate sampled goal states based using our reward metric defined in Section~\ref{sec:watchandmove}. Finally, we provide an oracle performance based on the optimal goal states generated by the oracle in the testing environments.

\begin{figure*}[t!]
\begin{center}
\centerline{\includegraphics[width=\textwidth]{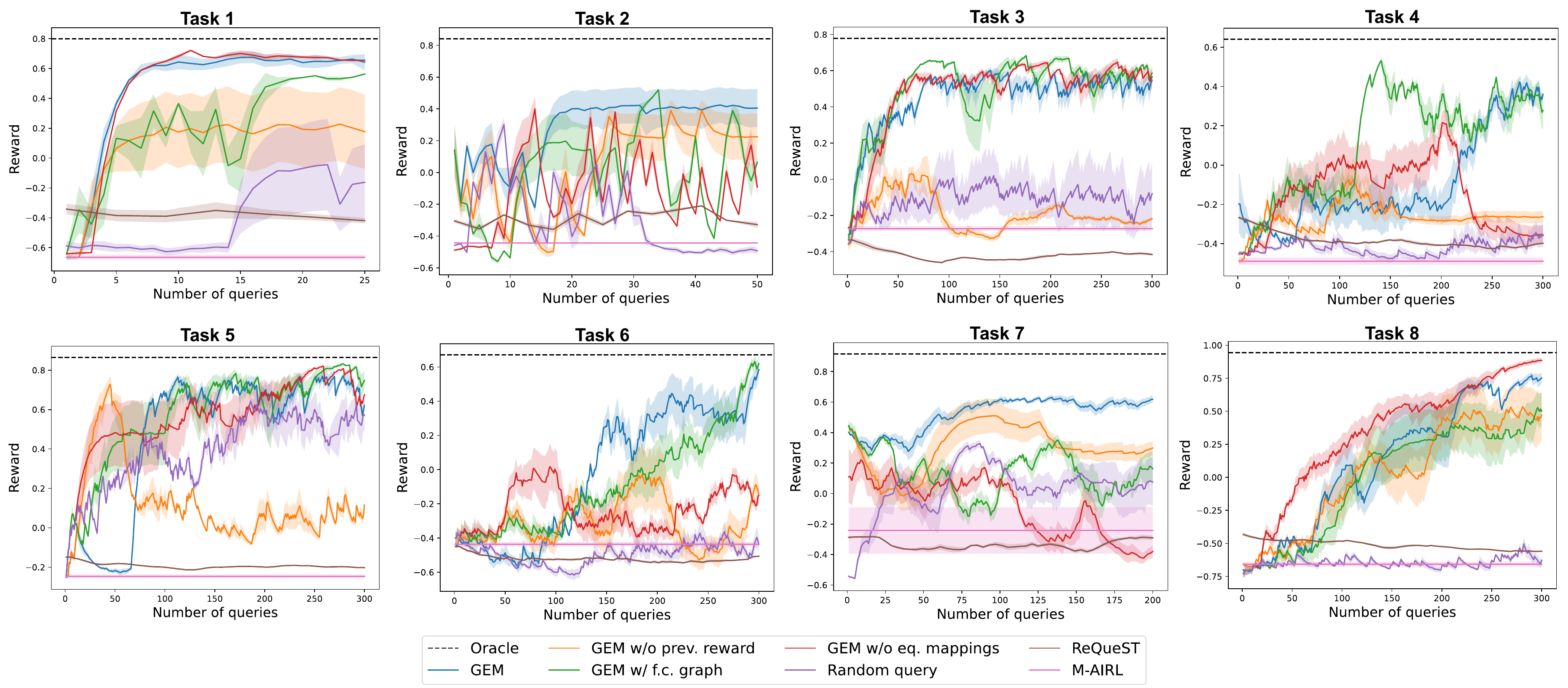}}
\caption{Testing performance of different methods trained with a simulated oracle on 8 Watch\&Move tasks (Task 1 to 8). We plot the reward metric in the testing environment using the learned model as a function of the number of queries. The dashed line indicates the reward for an optimal plan generated by the oracle. Note here the M-AIRL baseline provides the initial rewards for GEM and all of its variants and is not updated with the queries. We ran each method using three random seeds and show the standard errors as the shaded areas.}
\label{fig:results_simulated}
\end{center}
\end{figure*}

\begin{figure}[t!]
\begin{center}
\centerline{\includegraphics[width=0.95\columnwidth]{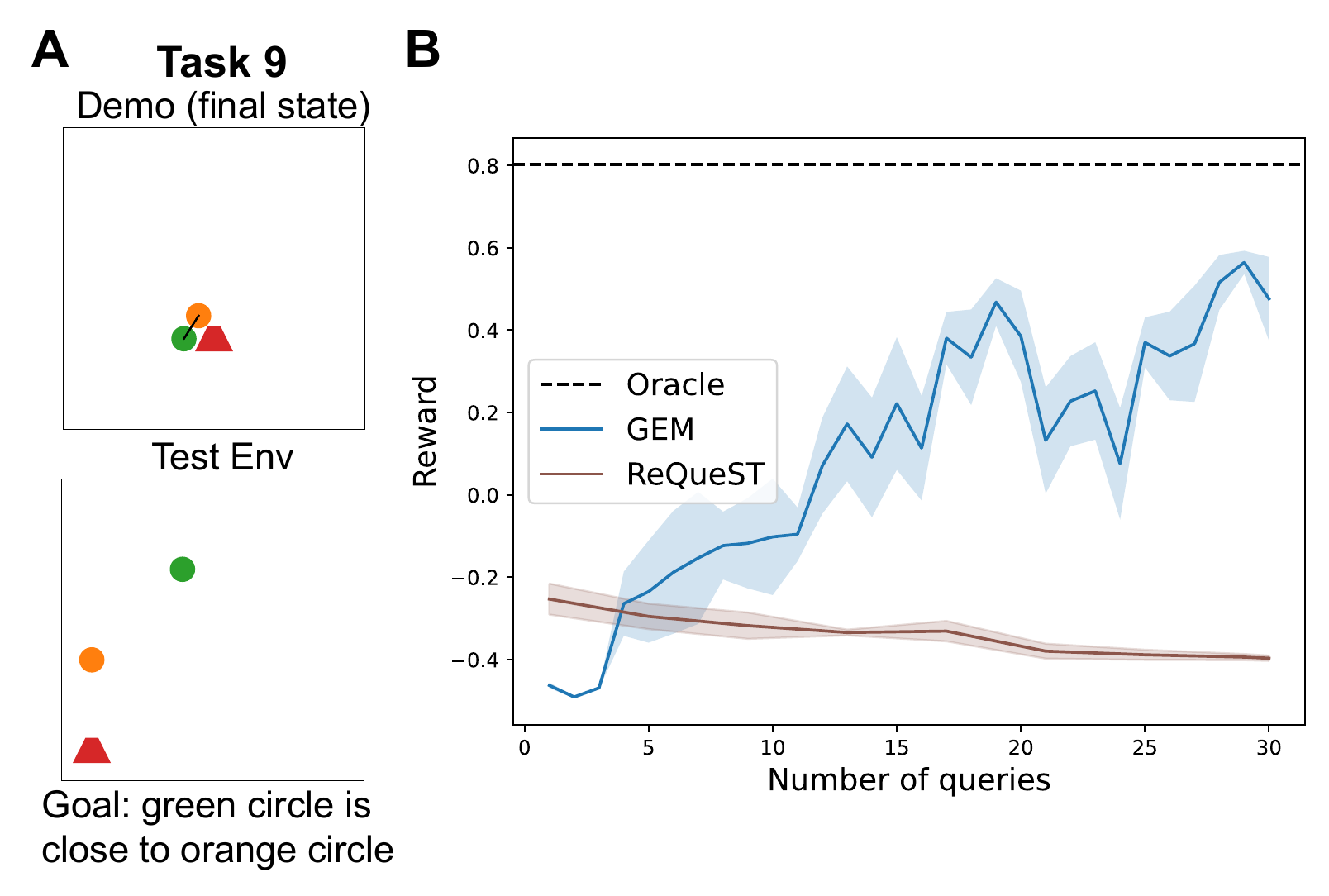}}
\caption{(\textbf{A}) Illustration of Task 9 used in the human experiment. (\textbf{B}) Testing performance of GEM and ReQueST trained with three human oracles for Task 9. The dashed line indicates the reward for an optimal plan generated by the oracle in the test environment. The shaded areas indicate the standard errors.}
\label{fig:results_human}
\end{center}
\vspace{-15pt}
\end{figure}

\subsection{Results with a Simulated Oracle}

We first conducted an evaluation on 8 Watch\&Move tasks with a simulated oracle that gives feedback for a query based on whether the state in the query satisfies the goal definition. We report the reward obtained in the testing environment using models trained with different methods with different numbers of queries in Figure~\ref{fig:results_simulated}. The results show that the initial reward trained by M-AIRL failed to reach the goal in the testing environment. However, with the active reward refinement enabled by GEM, the reward function was greatly improved. We also found that the graphs and the equivalence mapping assignments inferred by GEM provided meaningful representations of the intended spatial relationships for the tasks (we visualize the inferred graphs and equivalence mapping assignment, and example queries in Appendix~\ref{app:more_results}). In comparison, ReQueST and the random query variant failed to learn a generalizable reward function for all tasks. Other ablated variants could achieve success in some tasks but not in all 8 tasks, which verified the importance of i) minimizing the previous reward for the query sampling and ii) the joint inference of the graph structure and the equivalence mapping assignment.

\subsection{Human Experiment}
To evaluate how well GEM can work with real human oracles, we conducted a human experiment on Task 9, where we recruited three human participants as oracles. The participants gave their consent, and the study was approved by an institutional review board.

At each trial, a participant was instructed to verify whether a query state generated by an AI agent satisfied the ground truth goal of Task 9 (see Figure~\ref{fig:results_human}). Each participant interacted with GEM and ReQueST once and provided feedback for $30$ queries for each algorithm. Each GEM query took about $1$ minute due to training. We plot the rewards in the testing environment in Figure~\ref{fig:results_human}, which demonstrates that the reward function trained by GEM reaches a good performance in the new environment within 30 iterations, significantly outperforming the reward function trained by ReQueST. GEM also correctly inferred the necessary edge (the pair of circles) and its corresponding invariance type (rotation-invariant).

\section{Conclusion}

We have proposed a reward learning algorithm, GEM, for performing one-shot generalization for an object rearrangement task. In GEM, spatial goal relationships are represented by a graph-based reward function and state equivalence mappings assigned to the edges of the graph. GEM infers the graph structure and the equivalence mapping assignment by combing an initial reward learning using inverse reinforcement learning and a sample efficient active reward refinement. For evaluation, we designed a 2D physics simulation environment, Watch\&Move, and compared GEM against strong baselines on multiple tasks. We also conducted a human experiment to verify whether GEM can work with humans. The experimental results show that GEM was able to discover meaningful spatial goal representations. It significantly outperformed baselines, achieving a better generalization in unseen testing environments as well as a greater sample efficiency. We believe that GEM is a step towards solving the extremely challenging problem of one-shot imitation learning. 

 The success of GEM in simulation shows potential for real-world applications. First, GEM achieves a much greater sample efficiency and generalizability compared with SOTA (ReQueST), making it possible to learn the reward with humans. Second, there are several ways to apply GEM to learn task specifications that involve more objects with a reasonable amount of queries. The bottleneck of the sample efficiency is the number of edges in a graph. Since real-world tasks typically require a sparse graph, we can use heuristics (e.g., an object is usually only related to neighboring objects) to directly remove unlikely edges without queries. We can also take a hierarchical approach where we first learn rewards for subgraphs (e.g., a dining set) and then apply GEM to learn the final reward (e.g., multiple sets for a party of four) where each node is a subgraph.
 Finally, since our reward representation and the learning algorithm only require generic object states (e.g., positions, orientations), it is possible to learn rewards in different state spaces for different domains (such as shapes in a 3D space rather than shapes in a 2D space). We intend to study the real-world applications of GEM in the future.

The present work has a few limitations. First, we are only focusing on pairwise relationships between objects and representing them through two types of invariances. In the future, we intend to learn higher-order relationships by enabling message passing in the graphs and introducing additional types of invariance. Second, for real-world applications, we need to improve proposal sampling so that it can infer the graph structure and the equivalence mapping assignment more efficiently for a large number of objects. We can potentially achieve this by using language instructions to guide the inference. For instance, a human user can provide a language description of the task in addition to the physical demonstration to help achieve a better understanding of the task.

\section*{Acknowledgements}
We would like to thank Xavier Puig for his assistance with the experiment in VirtualHome. This work was supported by the DARPA Machine Common Sense program and ONR MURI N00014-13-1-0333.

\bibliography{example_paper}
\bibliographystyle{icml2022}

\newpage
\appendix
\onecolumn
\section{Implementation Details}\label{appA:implementation-details}

\subsection{State Equivalence Mappings}\label{sec:app_mappings}

\noindent\textbf{Applying a mapping to an edge once}. When applying a mapping to an edge, we randomly sample a variable necessary for the mapping. For the the rotation-invariant mapping, $\phi_1(x_1, x_2)$, we sample a random angle $d \sim \text{Uniform}(-\pi, \pi)$, and rotate $i$ around $j$ by the angle of $d$. For the scale-invariant mapping $\phi_2$, we randomly sample a scale $\rho \sim \text{Uniform}(0.1, 10.0)$, and move $i$ so that its distance from $j$ changes by the scale of $\rho$ while the relative orientation from $i$ to $j$ remains the same.  

\noindent\textbf{State transformation given all mappings}. For each graph, there is a corresponding mapping assignment for all edges $I=\{\delta_{ijk}\})_{(i, j) \in E, k\in K}$. We apply a mapping $\phi_k$ to edge $(i,j)$ when and only when $\delta_{ijk}=1$. If multiple mappings are assigned to an edge, they will be applied recursively. E.g., if $\phi_1$ and $\phi_2$ are assigned to $(i, j)$, the final state transformation for this edge is $\phi_2(\phi_1(x_i, x_j))$. Let $\tilde{x}_{ij}$ be the transformed edge, then the final transformed state is $\Phi(s, I) = \{\tilde{x}_{ij}\}_{(i,j) \in E}$, and the corresponding reward becomes:
\begin{equation}
    R(\Phi(s, I)|G) = \frac{1}{|E|}\sum_{(i,j)\in E} r(\tilde{x}_{ij}). 
\end{equation}
Note that since the state transformations are applied to each edge independently, any change in one edge will not affect other edges. During reward finetuning, we apply state transformation based on the proposed mapping assignment to each batch, so that the trained reward function reflects the intended invariance represented in the assigned equivalence mappings for each edge.

Additionally, we assume that the absolute coordinates of the objects do not matter in the goal specifications. Therefore, we also apply a random shift to all objects' coordinates after applying the state equivalence mappings (i.e., moving all objects together without changing their relative positions).

\subsection{Proposal Sampling}\label{sec:app_proposal}

Each proposal consists of a new graph and a new equivalence mapping assignment. Therefore, there are two general types of proposal sampling -- (1) graph sampling and (2) equivalence mapping assignment sampling. We use a prior probability $q_\text{type}$ to decide the type of sampling for each iteration. At each iteration, we first sample $u \sim \text{Uniform}(0, 1)$. If $u < q_\text{type}$, we choose to sample a new equivalence mapping assignment; otherwise, we sample a new graph.  For 3-object tasks, we use $q_\text{type} = 0.2$; for 4-object tasks, we use $q_\text{type} = 0.5$.

To sample a new graph, we can either add an edge or remove an edge. We define the chance of removing an edge as $q_\text{remove}$. Then we sample $u \sim \text{Uniform}(0, 1)$. When $u < q_\text{remove}$, we sample one of the edges from $G^{l-1}$ to remove; otherwise, we randomly add an edge that does not exist in $G^{l-1}$. Note that we consider undirected graphs in this work; we also avoid removing all edges to ensure a valid graph-based reward function. For all tasks, we use $q_\text{remove} = 0.5$.

To sample a new equivalence mapping assignment, we uniformly sample an edge $(i, j) \in E$ and a type of mapping $k \in K$, and change the corresponding assignment, i.e., $\delta_{ijk}^\prime = 1 - \delta_{ijk}^{l-1}$.

\subsection{Network Architecture}
The discriminator reward and value networks are implemented by a graph-based architecture, as opposed to the MLP architecture used in the original AIRL version. The input is the observation representation $s$ and graph $G$. We construct edge representations by concatenating every pair of object representations in the observation. We then apply 4 x (fully connected layers + ReLU) to each edge representation. We apply a final fully connected layer to each edge embedding to output a single value for each edge. The final reward is the average edge value \textit{only} for edges in $G$. 

\subsection{Training Details and Hyperparameters}

\textbf{Model-based state-based AIRL}: We build upon an AIRL implementation \cite{wang2020imitation} and add a model-based generator. For each task, M-AIRL is executed for 500k generator steps, the expert batch size is the length of the expert demonstrations. For the model-based policy, we set $\beta=0.3$ in Eq.~(\ref{eq:model-based-policy}). The discriminator is updated for 4 steps after every model-based generator execution. 
The model-based generator samples approximately 2k steps on each iteration.

\textbf{Query reward finetuning}:  We apply 5k network updates per query iteration. For optimizing the network, we use Adam optimizer \cite{kingma2014adam} with a learning rate of 0.0003. For each update, we sample a batch of 16 states for the regression loss and a batch of 16 pairs of positive and negative states for the reward ranking loss.

\textbf{Query selection}: we set $\lambda=0.2$ in Eq~\ref{eq:query_gen}.

\textbf{Random query variant}: we always assume a fully connected graph and do not assign equivalence mappings to the edges. We collect the positive and negative sets and refine the reward function using the same loss function defined in Eq.~(\ref{eq:optimization}).

\subsection{Sampling Goal States in Testing}
During testing, we sample a goal state, $s^*=(x_i^*)_{i\in N}$, by jointly maximizing the learned reward and minimizing the displacement. I.e.,

\begin{equation}
s^* = \argmax_{s=(x_i)_{i\in N}} R(s, G) - 0.02 \sum_{i\in N} ||x^0_i - x_i||_2.
\end{equation}

For the experiments using the simulated oracle, we use the reward corresponding to the sparsest graph accepted that has been accepted up until the current iteration. For the human experiments, we directly use the reward accepted at each iteration so that the results are more resilient to the noise in participants' responses.

Since this work focuses on learning goals, we directly evaluate feasible sampled goal states in the experiments. However, it is also possible to evaluate an episode using a planner to reach the sampled goal states.

\subsection{Watch\&Move Environment}\label{sec:app_watchandmove}

\begin{figure*}[t!]
\begin{center}
\centerline{\includegraphics[width=\textwidth]{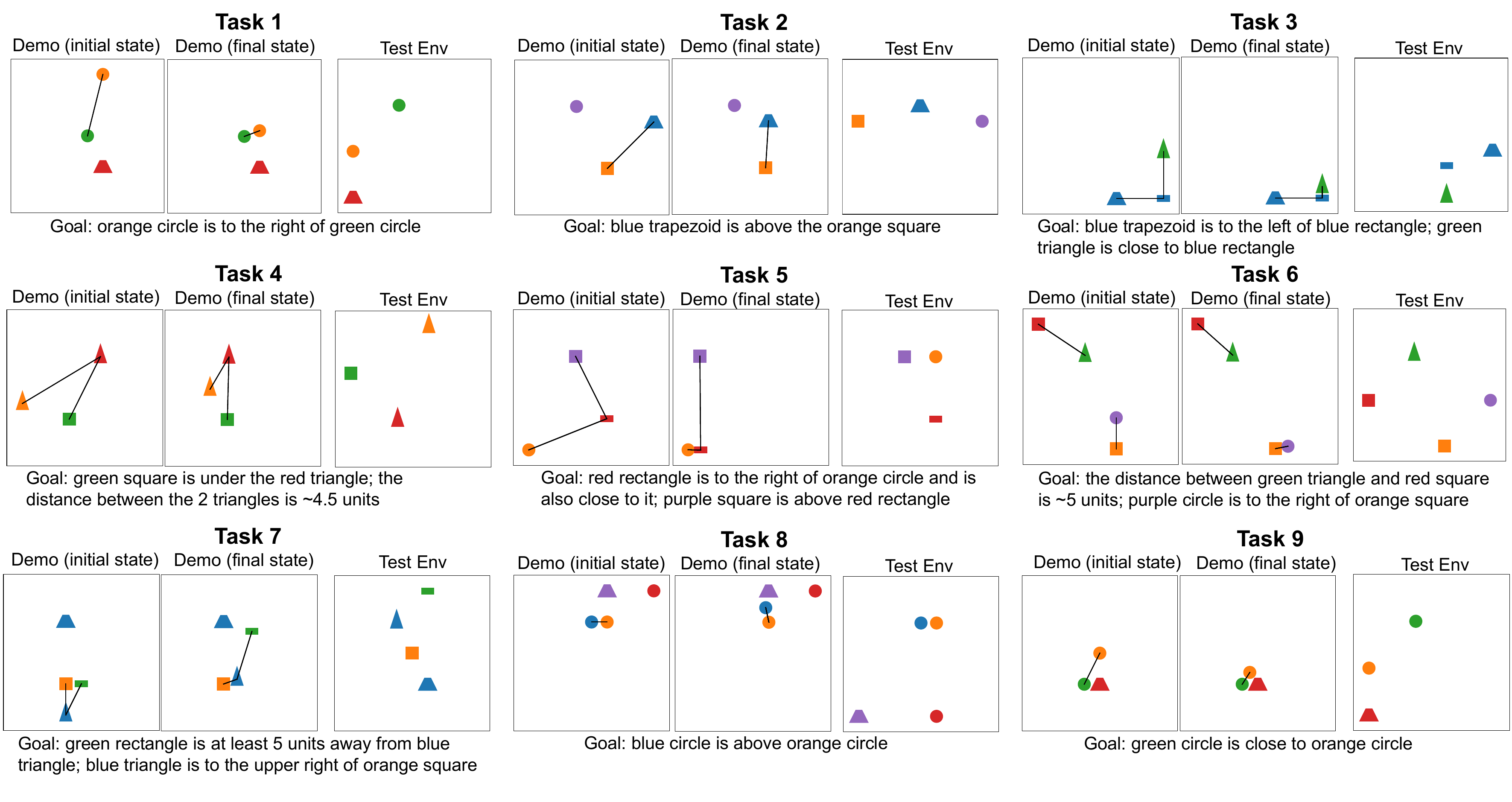}}
\caption{Illustration of all 9 Watch\&Move tasks. For each task, we show the demonstration, the testing environment, and the goal definition. We also show the goal relationships by visualizing the corresponding edges.}
\label{fig:watchandmove_tasks}
\end{center}
\end{figure*}

Figure~\ref{fig:watchandmove_tasks} illustrates the setup (the expert demonstration, the test environment, and the goal description) for each Watch\&Move task. The demonstrations of these tasks present various sources of confusion. For instance, there can be irrelevant objects (e.g.,the purple trapezoid and the red circle in Task 8) or goal objects that are never moved in the demonstration due to their initial states being satisfactory (e.g., the blue trapezoid and the blue rectangle in Task 3). Expert demonstrations were created with a planner introduced in \cite{NetanyahuPHASE2021}, with a length ranging from 8 to 35 steps.

The state space in Watch\&Move is represented by $s\in\mathbb{R}^{N\times 13}$ where $N$ is the number of objects in the environment. $13$ dimensions are composed of the coordinate of the object center, the object's angle, and one hot encodings of the object's shape and color.
The action space is discrete, containing 11 possible actions per object (8 directions, turning right and left and stopping), and the object id. The action space is proportional to the number of objects. We use PyBox2D\footnote{https://github.com/pybox2d/pybox2d} to simulate the physical dynamics in the environment.

The goal relationships used to create Watch\&Move are specified in Table \ref{tab:predicate-def}. These could be easily extended to any pairwise spatial relation, such as touching, covering, no contact, orientation, etc.

\begin{table}[t]
\caption{Definition of goal relationships in Watch\&Move tasks. For reference, the radius of a circle is $0.8$ units.}
\label{tab:predicate-def}
\vskip 0.15in
\begin{center}
\begin{small}
\begin{sc}
\begin{tabular}{rl}
\toprule
Pairwise Goal Relationship & Definition\\
\midrule
close [\cite{srivastava2022behavior}] & 
objects distance $<$ $2.5$ units \\
Left/Right [\cite{johnson2017clevr, yan2020robotic}] & angle between objects and x axis $<$ $0.1\pi$ \\
Above/Below [\cite{yan2020robotic}] & $0.4\pi$ $<$ angle between objects and y axis $<$ $0.6\pi$\\
Diagonal & both coordinates of one object $>$ other's\\
Distance x & objects distance within 0.5 unit buffer around x\\
At least within distance x & objects distance $>$ x \\
\bottomrule
\end{tabular}
\end{sc}
\end{small}
\end{center}
\vskip -0.1in
\end{table}

\subsection{ReQueST}\label{app:request}
ReQueST \cite{reddy2020learning} is a method for estimating a reward ensemble that can safely generalize to environments with different initial state distributions. ReQueST generates queries from a generative model that optimizes four objectives. Each state in each query receives accepted or rejected feedback from an oracle and is used as a positive or negative example in reward training. To ensure a fair comparison with our approach, we implement ReQueST with the following changes. 
\begin{itemize}
\item The original method does not use an expert demonstration, therefore we pre-train the ensemble reward functions with the expert demonstration.
\item The reward architecture is similar to ours, where the final edge embeddings are sum-pooled and fed to a fully connected layer followed by softmax for the classification. Note that here we only have two classes -- neutral and good.
\item The original paper learns a world model from random sampling in the environment. We use the ground truth world model provided by the physics simulation, similar to GEM.
\item Each query is of length 1 (as in the pointmass environment in ReQueST) sampled similarly to the AIRL generator sampling in GEM. Starting from the final state of the expert demonstrations, each query is sampled relative to the last sampled point, matching with the query generation procedure in GEM.
\end{itemize}

We use an ensemble of 4 rewards (as in the pointmass environment in ReQueST), and train them using the Adam optimizer with a learning rate of 0.0003. We have 1k pre-training steps and 10 training steps per every 4 queries.

\begin{figure*}[t!]
\begin{center}
\centerline{\includegraphics[width=\textwidth]{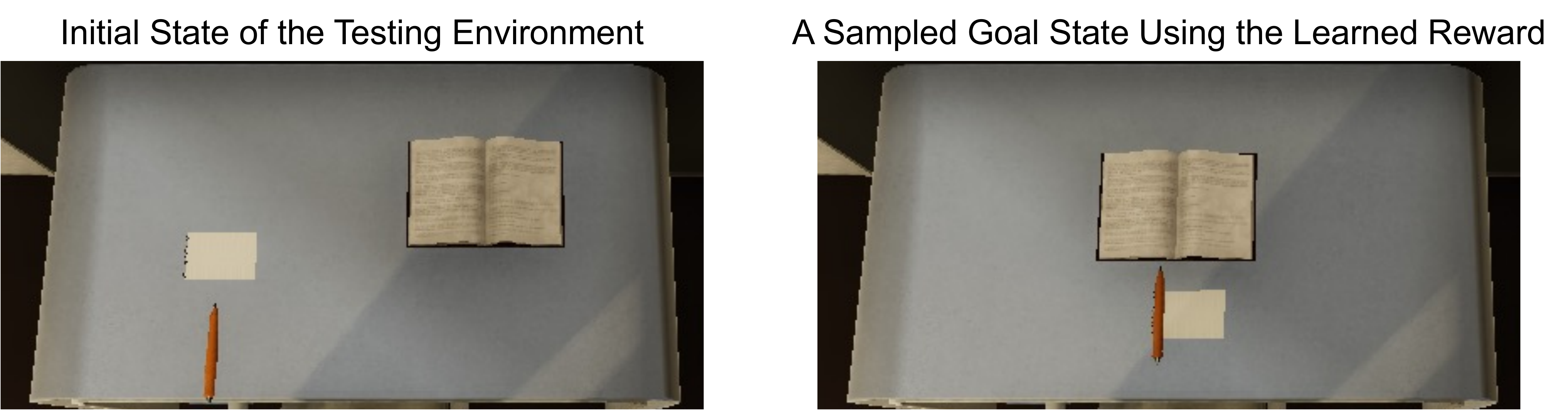}}
\caption{A real-world task simulated in VirtualHome that has the same goal specification as Task 5. Specifically, the task is to set up a desk for a left-handed person to take notes while reading a book. We visualize the testing environment and the sampled goal state  for this real-world task using the reward function learned by GEM for Task 5. The sampled goal state satisfies the true goal specification while minimizing the displacement of the objects.}
\label{fig:vh}
\end{center}
\end{figure*}

\section{Additional Results}\label{app:more_results}

\subsection{Evaluation in a Real-World Task}\label{sec:app_vh}

To show the connection between Watch\&Move tasks and real-world tasks, we use VirtualHome \citep{puig2018virtualhome,puig2021watchandhelp}, a realistic virtual household simulation platform, to instantiate Task 5 in a real work setting -- setting up a desk for a left-handed person, i.e., a book in the front of a notepad and a pen on the left of and close to the notepad. In Figure~\ref{fig:vh}, we show the reward trained by GEM on the original Task 5 can also achieve success in this real-world task which shares the same ground truth goal with Task 5 but requires the rearrangement of real-world objects.

\subsection{Learning Results}

\begin{figure*}[t!]
\begin{center}
\centerline{\includegraphics[width=0.65\textwidth]{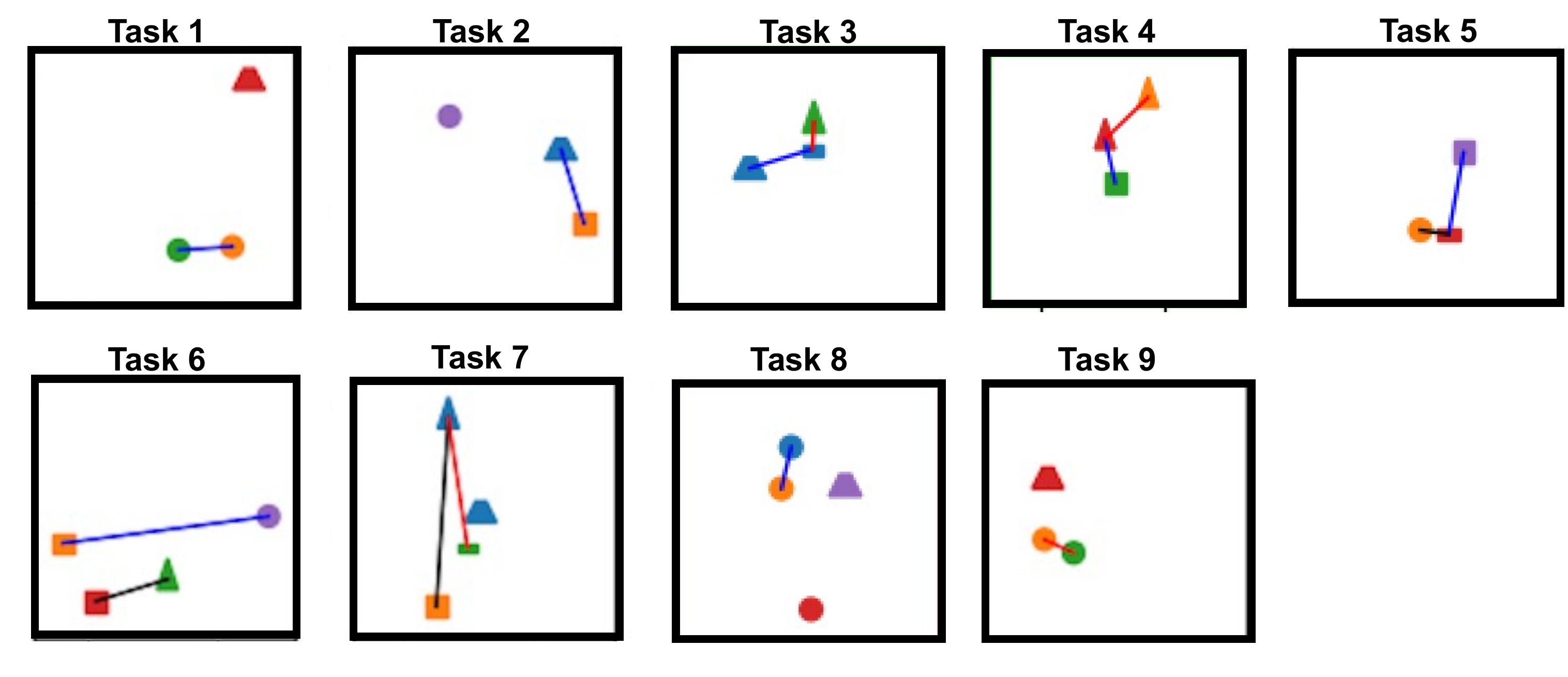}}
\caption{Illustration of the inferred graphs and equivalence mapping assignments as well as the corresponding query states that lead the proposal acceptance for all 9 tasks (from one of the three runs). The colors of the edges indicate the assigned mappings. Black: no mapping is assigned; red: the rotation-invariant mapping is assigned; blue: the scale-invariant mapping is assigned; purple: both the rotation-invariant and the scale invariant mappings are assigned. }
\label{fig:hypotheses}
\end{center}
\end{figure*}

We show the inferred graphs and the equivalence mapping assignments for all 9 tasks in Figure~\ref{fig:hypotheses}. The inferred graphs correctly identified the goal relationships. The assigned equivalence mappings also revealed the invariance properties of the intended spatial relationships in most cases. 

\begin{figure*}[t!]
\begin{center}
\centerline{\includegraphics[width=0.9\textwidth]{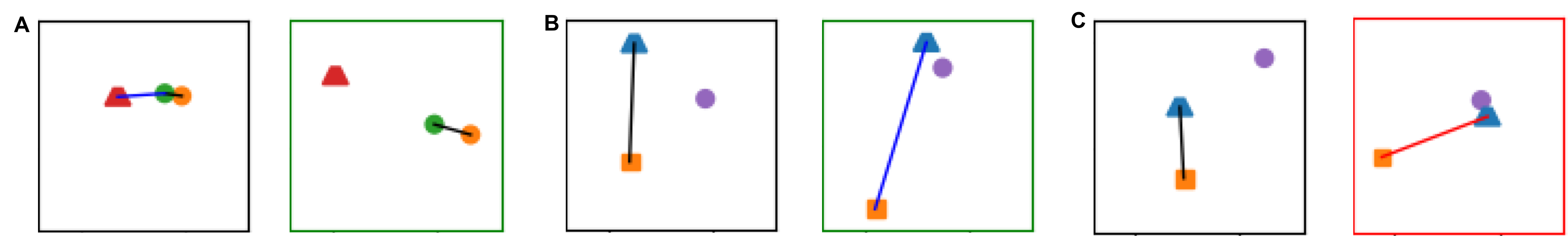}}
\caption{(\textbf{A})-(\textbf{C}) are example queries showing the effect of the new graph and equivalence mapping assignment. Each example first shows the state and proposal before the query, and then shows the new proposal and the sampled query state. The colors of the edges indicate the assigned mappings. Black: no mapping is assigned; red: the rotation-invariant mapping is assigned; blue: the scale-invariant mapping is assigned; purple: both the rotation-invariant and the scale-invariant mappings are assigned. The colors of the boxes indicate oracle feedback (red: reject, green: accept). In (\textbf{A}), an edge was removed, and the irrelevant object was consequently placed far away from the remaining objects. The example in (\textbf{B}) shows that when the scale-invariant mapping was assigned to an edge, the sampled query changed the distance between the objects connected by that edge while generally preserving the relative orientation between the two. On the other hand, when the rotation-invariant mapping was assigned as in (\textbf{C}), one of the objects was rotated around the other connected object and there was little change in the distance between the two objects.}
\label{fig:example_queries}
\end{center}
\end{figure*}

Figure~\ref{fig:example_queries} depicts typical queries that are sampled by GEM to verify the proposed graphs and the equivalence mapping assignment. The examples here demonstrate how the informative queries were able to help differentiate two different reward functions.

\begin{figure*}[t!]
\begin{center}
\centerline{\includegraphics[width=\textwidth]{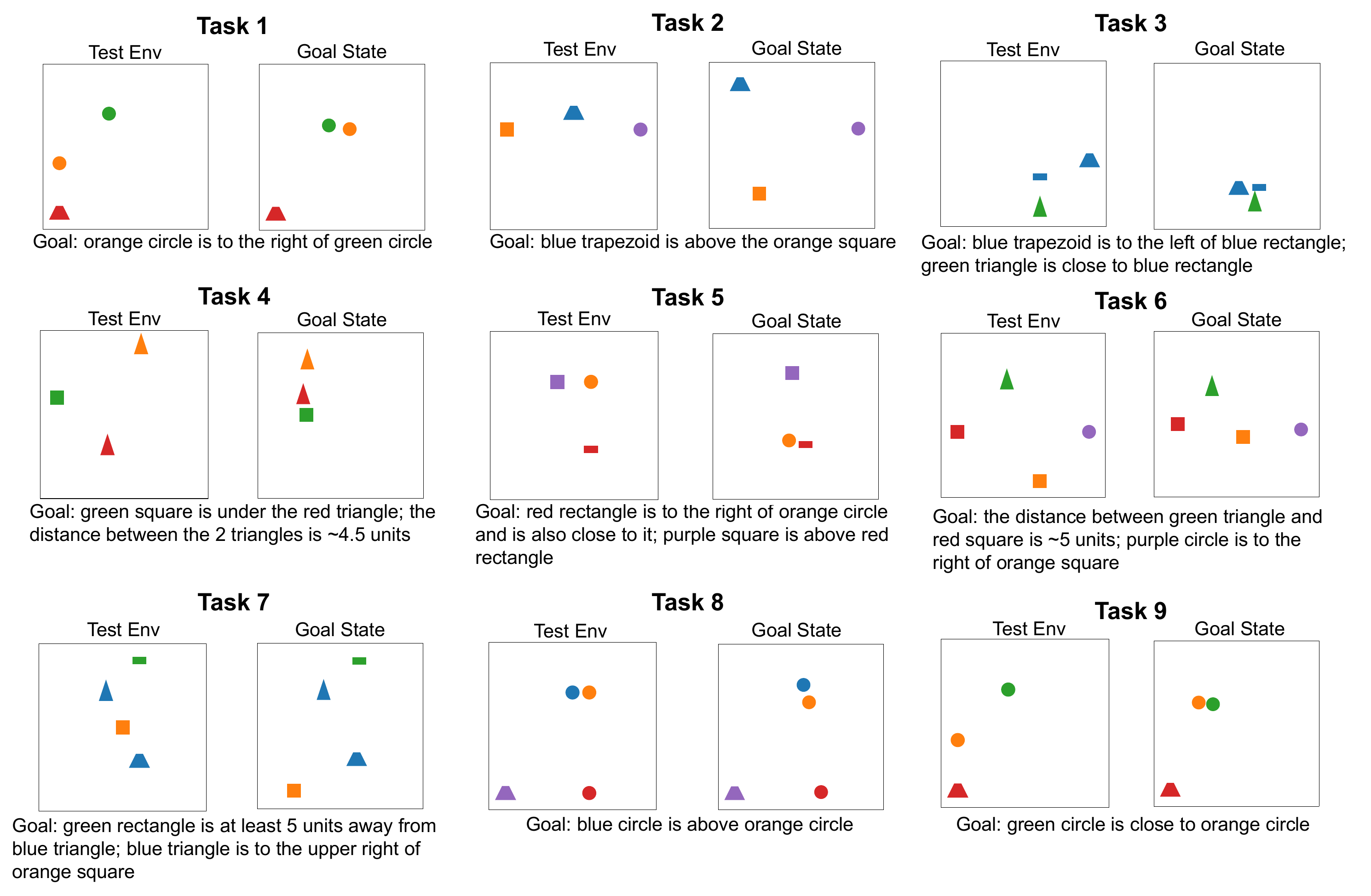}}
\caption{The testing environments and the sampled goal states based on the reward functions learned by GEM for the corresponding tasks. }
\label{fig:sampled_goal_states}
\end{center}
\end{figure*}

\subsection{Qualitative Results}

Figure~\ref{fig:sampled_goal_states} shows the sampled goal states using the reward functions learned by GEM. These goal states not only satisfy the goal specifications, but are also efficient (small displacement compared to the initial states in the testing environments). This demonstrates that the sampled goal states are not simply copying the final state of the expert demonstration but indeed reflect the intended spatial relationships.

\end{document}